\documentclass[11pt,a4paper]{article}
\usepackage{times}
\usepackage{latexsym}
\usepackage{xcolor}

\usepackage{multirow}
\usepackage{booktabs}
\usepackage{graphicx}
\usepackage{amsmath}
\usepackage{amssymb}
\usepackage[ruled, lined, linesnumbered, commentsnumbered, longend]{algorithm2e}
\usepackage{booktabs,tabulary}
\usepackage{siunitx}
\usepackage{threeparttable}
\usepackage{tabularx}
\usepackage{enumitem}
\usepackage{nccmath}
\usepackage{bm}
\PassOptionsToPackage{breaklinks}{hyperref}
\usepackage{xurl}
\usepackage[hyperref]{emnlp2020}

\newcommand{\norm}[1]{\left\lVert#1\right\rVert}

\usepackage{microtype}

\aclfinalcopy 
\def\aclpaperid{3562} 

\title{Extremely Low Bit Transformer Quantization \\ for On-Device Neural Machine Translation}

\author{
Insoo Chung\thanks{\hspace{0.2cm}Equal Contribution.} \hspace{1cm} Byeongwook Kim\footnotemark[1] \hspace{1cm}  Yoonjung Choi \hspace{1cm} Se Jung Kwon\\ 
\textbf{Yongkweon Jeon \hspace{1.0cm} Baeseong Park \hspace{1.0cm} Sangha Kim \hspace{1.0cm} Dongsoo Lee}\\
Samsung Research, Seoul, Republic of Korea\\
\texttt{\{insooo.chung, byeonguk.kim, yj0807.choi, sejung0.kwon,}\\ 
\texttt{dragwon.jeon, bpbs.park, sangha01.kim, dongsoo3.lee\}@samsung.com}\\
}

\date{}

\begin{document}
\maketitle
\begin{abstract}

The deployment of widely used Transformer architecture is challenging because of heavy computation load and memory overhead during inference, especially when the target device is limited in computational resources such as mobile or edge devices. Quantization is an effective technique to address such challenges. Our analysis shows that for a given number of quantization bits, each block of Transformer contributes to translation quality and inference computations in different manners. Moreover, even inside an embedding block, each word presents vastly different contributions. Correspondingly, we propose a mixed precision quantization strategy to represent Transformer weights by an extremely low number of bits (e.g., under 3 bits). For example, for each word in an embedding block, we assign different quantization bits based on statistical property. Our quantized Transformer model achieves 11.8$\times$ smaller model size than the baseline model, with less than -0.5 BLEU. We achieve 8.3$\times$ reduction in run-time memory footprints and 3.5$\times$ speed up (Galaxy N10+) such that our proposed compression strategy enables efficient implementation for on-device NMT.

\end{abstract}

\section{Introduction}
\label{sec:introduction}

Transformer \cite{transformer} is one of the state-of-the-art approaches for Neural Machine Translation (NMT), and hence, being widely accepted. For example, in WMT19 machine translation tasks, it is reported that 80\% of submitted systems have adopted the Transformer architecture \cite{wmt2019}. Note that high translation quality of Transformer models entails a large number of parameters. Moreover, the Transformer model is inherently much slower than conventional machine translation approaches (e.g., statistical approaches) mainly due to the auto-regressive inference scheme \cite{auto-regressive} incrementally generating each token. As a result, deploying the Transformer model to mobile devices with limited resources involves numerous practical implementation issues.

To address such implementation challenges with little degradation in translation quality, we study a low-bit quantization strategy for Transformer to accomplish high-performance on-device NMT. We note that most previous studies to compress Transformer models utilize uniform quantization (e.g. INT8 or INT4). While uniform quantization may be effective for memory footprint savings, it would face various issues to improve inference time and to maintain reasonable BLEU score. For example, even integer arithmetic units for inference operations present limited speed up \cite{intel8bit} and resulting BLEU score of quantized Transformer can be substantially degraded with low-bit quantization such as INT4 \cite{fully-quantized-transformer}.

While determining the number of quantization bits for Transformer, it is crucial to consider that each component of Transformer may exhibit varied sensitivity of quantization error toward degradation in translation quality \cite{qbert}. Accordingly, a mixed precision quantization can be suggested as an effort to assign different numbers of quantization bits depending on how each component after quantization is sensitive to the loss function. In addition, as we illustrate later, even assigning different quantization bits for each row of an embedding block can further reduce the overall number of quantization bits of the entire Transformer model. Our proposed quantization strategy, thus, provides a finer-grained mixed precision approach compared to previous methods, such as \cite{hawq, cnn-mixed-quantization, adaptive-quant, qbert} that consider layer-wise or matrix-wise mixed precision.


Accommodating distinguished implementation properties (e.g., latency and translation quality drop) of each component in Transformer, we propose the following methodologies to decide precision of a block: 1) in the case of embedding block, statistical importance of each word is taken into account and 2) for encoder and decoder blocks, sensitivity of each quantized sub-layer is considered. The main contributions of this paper are as follows:

\begin{itemize}[noitemsep]
    \item We propose a mixed precision quantization strategy while embedding block allows another level of mixed precision in word level according to statistical properties of natural language.
    \item Our proposed quantization scheme allows the number of quantization bits to be as low as under 3 bits for the Transformer with little BLEU score degradation (under -0.5 BLEU).
    \item We demonstrate that our quantization technique reduces a significant amount of run-time memory and enhances inference speed so as to enable fast on-device machine translation by large Transformer models.
\end{itemize}

\section{Background}
\label{sec:background}

\subsection{Transformer}
\label{sec:transformer}

Transformer adopts an an encoder-decoder architecture \cite{encoder-decoder-structure} composed of three different blocks: encoder, decoder and embedding that account for 31.0\%, 41.4\%, and 27.6\%, respectively, in terms of the number of parameters in a Transformer base model. An embedding block is a single weight matrix that serves multiple purposes in the Transformer. For example, each row in the embedding block represents a word in a bi-lingual vocabulary. Another purpose of the embedding block is to serve as a linear transformation layer which converts decoder outputs to next token probabilities as suggested in \citet{shared-embedding}. Encoder and decoder blocks are composed of multiple layers while each layer employs attention and feed-forward sub-layers.

Due to auto-regressive operations during inference of Transformer \cite{auto-regressive}, the correlation between the number of operations and the number of parameters can be vastly different for each component. Based on such different correlations, Transformer's inference scheme can be divided into encoding steps of high parallelism and decoding steps of low parallelism. As for encoding steps, given a sequence in the source language, a single forward propagation of the encoder produces a sequence of hidden representations for all words in a given sequence. In each decoding step, decoder and embedding blocks produce a probability distribution of possible words, \textit{one word at a time}. Unlike encoding steps, the computation of decoding steps is not parallelizable because each decoding step depends on outputs of all prior decoding steps.




Note that such lack of parallelism during decoding steps potentially induces the memory wall problem in practice with commodity hardware; parameters of decoder and embedding blocks are required to be loaded to cache and unloaded from the cache repeatedly throughout decoding steps. Furthermore, an embedding block is usually represented by a significantly large matrix that also incurs the memory wall problem \cite{biqgemm}.

\subsection{Non-uniform Quantization Based on Binary-codes}
\label{sec:bcq}

Quantization approximates full precision parameters in neural networks by using a small number of bits \cite{vector-quant, binary-weight-network, greedy-binary-code-quant, tf-lite}. One of widely adopted quantization methods is uniform quantization.
Uniform quantization performs mapping of full precision parameters into one of $2^q$ values ranging from $0$ to $2^{q}{-}1$ that correspond to a range between the minimum and the maximum full precision parameters, where $q$ denotes the number of quantization bits. Lower precision can reduce the computation cost of arithmetic operation such as multiplication and addition only if all inputs to arithmetic operations (i.e., activations) are also quantized \cite{tf-lite}. Furthermore, high quantization error may occur when a parameter distribution involves extreme outliers \cite{quant-outlier}.

As such, non-uniform quantization methods are being actively studied to better preserve expected value of parameters which is critical to maintaining model accuracy \cite{binary-connect}. By large, non-uniform quantization methods include codebook-based quantization and binary-code based quantization. Even though codebook-based quantization reduces off-chip memory footprint, computational complexity is not reduced at all because of mandatory dequantization procedure during inference \cite{and-the-bit-goes-down, survey-quant}. On the other hand, quantization based on binary-code (${\in}\{-1, +1\}$) can achieve both high compression ratio and efficient computation \cite{ binary-weight-network, greedy-binary-code-quant, amq, biqgemm}.

\begin{figure}[!t]
  \includegraphics[width=\linewidth]{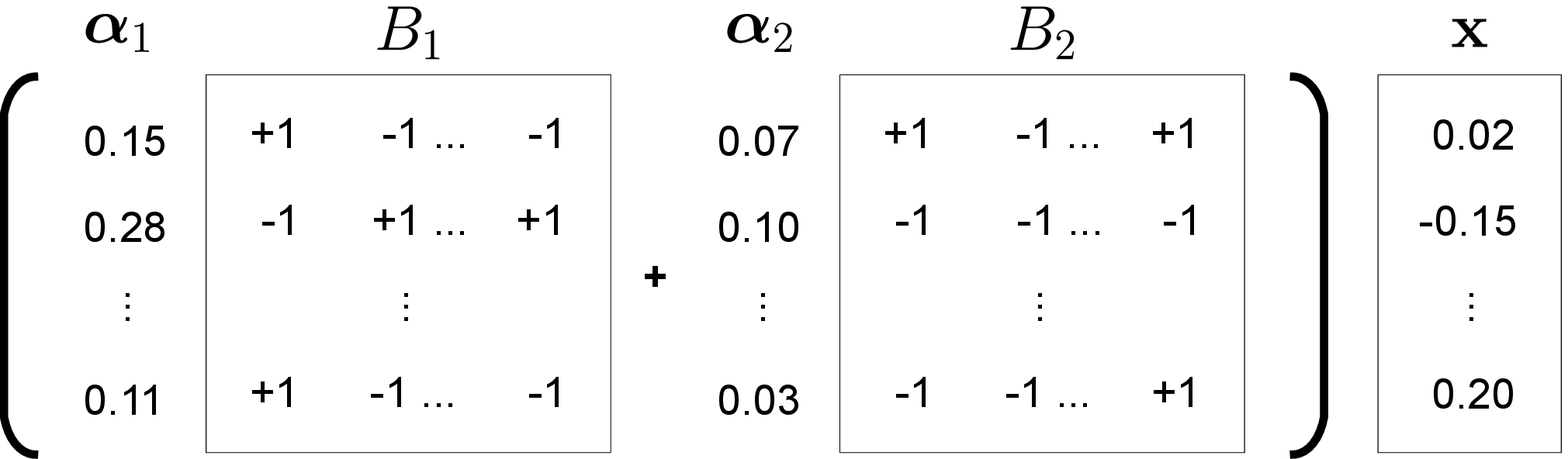}
  \caption{
  In 2-bit binary-code based quantization, each row of $W$ is approximated to 2 sets of binary code weights ($\{B_1,B_2\}$) and 2 vectors of full precision scales ($\{\boldsymbol{\alpha}_1,\boldsymbol{\alpha}_2\}$).}
  \label{fig:bcq-format}
\end{figure}

In this paper, we adopt non-uniform binary-code based quantization as our method of quantization. Non-uniform quantization based on binary-code maps a full precision vector $\mathbf{w} {\in} \mathbb{R}^p$ to a scaling factor $\alpha_i {\in} \mathbb{R}$, and a binary vector $\mathbf{b}_i {\in} \{-1, +1\}^p$, where $(1 {\leq} i {\leq} q)$. Note that $p$ is the length of a vector and $q$ denotes the number of quantization bits. Then, $\mathbf{w}$ is approximated as $\sum_{i=1}^q\alpha_{i}\mathbf{b}_i$. Scaling factors and binary vectors are obtained as follows:

\begin{equation}
\label{eq:bcq-quant}
arg\min_{\alpha_i, \mathbf{b}_i} \norm{\mathbf{w}-\sum_{i=1}^q\alpha_{i}\mathbf{b}_i}^2
\end{equation}

To minimize the quantization error formulated in Eq.~\ref{eq:bcq-quant}, heuristic approaches have been proposed \cite{greedy-binary-code-quant, amq}. 

For matrix quantization, the binary-code based quantization can be simply applied to each row or column of a matrix. With a matrix quantized into binary matrices $\{B_1,B_2,...,B_q\}$ and scaling factor vectors $\{\boldsymbol{\alpha}_1, \boldsymbol{\alpha}_2, ... ,\boldsymbol{\alpha}_q\}$, the matrix multiplication with full precision vector $\mathbf{x}$ produces an output vector $\mathbf{y}$ as follows:

\begin{equation}
\label{eq:bcq-matmul}
\mathbf{y} = \sum_{i=1}^{q}(\boldsymbol{\alpha}_i \circ (B_i\cdot\mathbf{x})),
\end{equation}
where the operation $\circ$ denotes element-wise multiplication. Figure \ref{fig:bcq-format} is an illustration of Eq.~\ref{eq:bcq-matmul}. Intermediate results of $B_i\cdot\mathbf{x}$ can be pre-computed for further compute-efficiency \cite{biqgemm}. This allows the efficient matrix multiplication of quantized Transformer weights and full precision activation.

\section{Quantization Strategy for Transformer}
\label{sec:quantization-strategy}

For Transformer, we suggest the following two techniques to decide the number of quantization bits for each block: 1) in the case of embedding block, frequency of each word is taken into account and 2) for encoder and decoder blocks, we find the minimum number of quantization bits for each type of sub-layers that allows reasonable degradation in BLEU score after quantization.

\subsection{Embedding}
\label{sec:embedding-quant}

\begin{figure}[!t]
    \centering
    \includegraphics[width=1.0\linewidth]{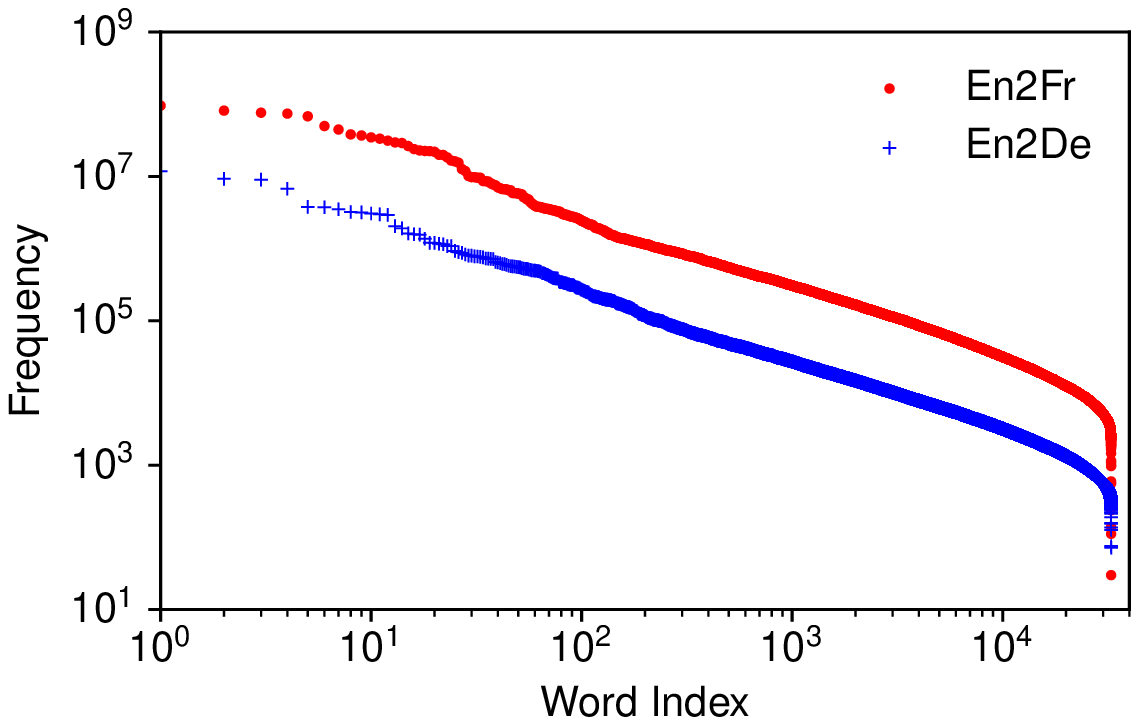}
    \caption{The distributions of word frequency in trainsets. Word indices range from 1 to 32768 where words are sorted in descending order of frequency.}
    \label{fig:word-freq}
\end{figure}

It has been reported that the word frequency distribution can be approximated as power-law distribution \cite{group-reduce}.
Such power-law distribution is illustrated in Figure \ref{fig:word-freq} that presents word frequency distribution in WMT14 datasets. Note that 1\% of word vectors account for around 95\% of word frequency for both En2Fr and En2De.
Intuitively, if word vectors are compressed by the same compression ratio, then word vectors with high frequency in a corpus would result in higher training loss after compression, compared to word vectors with low frequency.
\citet{group-reduce} utilizes frequency to provide different compression ratios in different groups of words using low-rank approximation.
To the best of our knowledge, word frequency has not yet been considered for Transformer quantization.

We assume that highly skewed word frequency distribution would lead to a wide distribution of the number of quantization bits per word.
In such a case, an embedding block may require a substantially high number of quantization bits that would be the maximum in the distribution of the number of quantization bits per word. For example, even though \citet{qbert} successfully quantized the parameters in attention and feed-forward sub-layers of the BERT architecture \cite{bert} into 2-4 bits, 8 bits or higher number of bits were used to represent a parameter in the embedding block.

\begin{algorithm}[h]
    \SetAlgoLined
    \SetKwInOut{KwIn}{Input}
    \SetKwInOut{KwOut}{Output}
    \KwIn{Embedding matrix $E$ of shape $[v, d_{model}]$; number of clusters $b$; the ratio factor $r$;}
    \KwOut{Quantized representation $\hat{E}$}
    Sort $E$ in descending order of word frequency \;
    $idx = 0$ \;
    \For{$i = 0 ... b-1$}{
        Compute number of word-vectors in $i$-th cluster, $c_{size}^i = \frac{v}{\sum_{k=0}^{b - 1}{r^{k}}} \cdot r^{i}$ \;\label{alg:line:r-define}
        Compute target bit-precision for $i$-th cluster, $c_{bit}^i = b - i$ \;\label{alg:line:b-define}
        \For{$j = 0 ... c_{size}^i$} {
            Initialize $w_{idx}$ = $idx$-th row of $E$ \;
            Quantize word vector $w$ to $c_{bit}^i$ bit, $\hat{w}_{idx} = quantize(w, c_{bit}^i)$ \;
            Increment $idx$ by 1 \;
        }
    }
    Output: $\hat{E} = \{\hat{w}_{0}, \hat{w}_{1}, ..., \hat{w}_{v-1}\}$
    \caption{\label{alg:embedding} Embedding quantization}
\end{algorithm}

The underlying principle to quantize embedding blocks is that the number of quantization bits for each word vector is proportional to the frequency in a corpus. To assign a low number of quantization bits to most of the words under such a principle, first, we group word vectors into clusters according to word frequency. $r$ acts as an exponential factor in deciding the number of word vectors in each cluster as in line \ref{alg:line:r-define} of Algorithm \ref{alg:embedding}. $b$ denotes the number of clusters and acts as a variable for quantization bits such as line \ref{alg:line:b-define} of Algorithm \ref{alg:embedding}.
For example, with $b{=}4$ and $r{=}2$, word vectors are clustered into clusters of ratio $r^0{:}r^1{:}r^2{:}r^3{=}1{:}2{:}4{:}8$, then assigned bits as much as $\{b, b{-}1, b{-}2, b{-}3\}=\{4, 3, 2, 1\}$. We empirically set $b{=}4$ for all of our embedding quantization experiments.

\begin{figure}[h]
    \centering
    \includegraphics[width=1.0\linewidth]{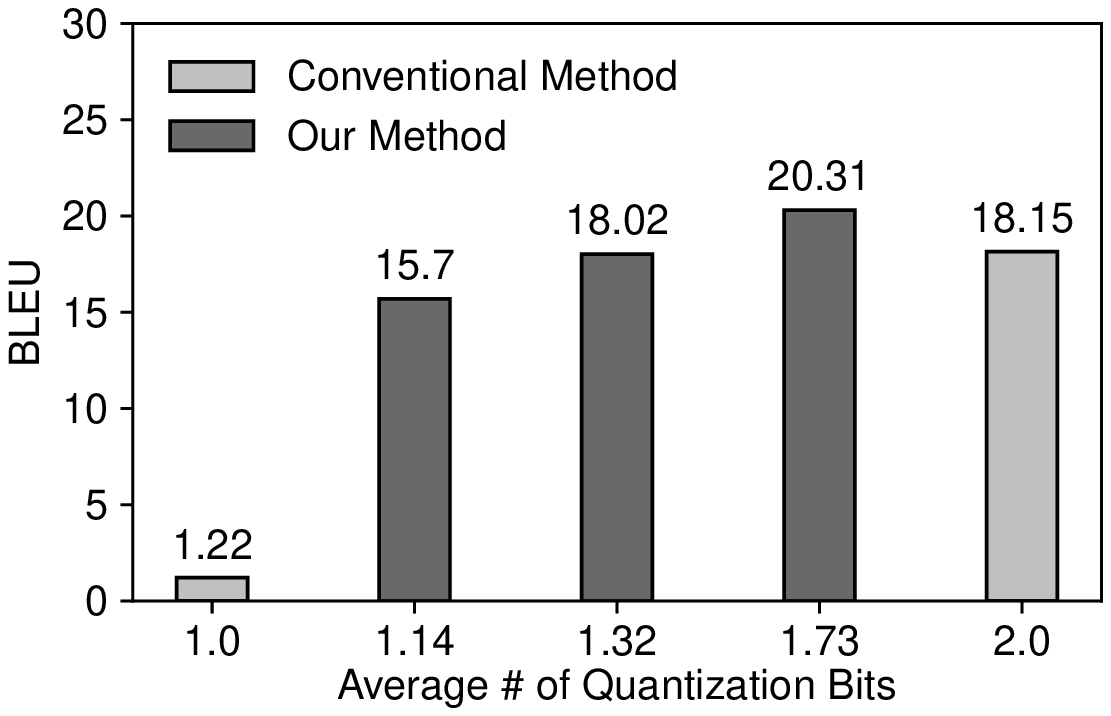}
    \caption{\label{fig:embed-datafree}Detokenized-BLEU (beam=4) on newstest2013 after quantizing the embedding block without retraining.}
\end{figure}

Figure \ref{fig:embed-datafree} shows our experimental results with $r{\in}\{2,4,8\}$. For $r{=}2$ , the average number of quantization bits in the embedding block is 1.73, and for $r{=}4$, it becomes 1.32. With our embedding quantization method, higher translation quality in terms of BLEU score can be achieved with lower number of quantization bits as compared to the conventional quantization methods that assign the same number of quantization bits to all word vectors. For example, the Transformer model with 1.73-bit quantized embedding produces more accurate translations than the model with conventional (fixed) 2-bit quantized embedding block.

Algorithm \ref{alg:embedding} assigns 1-bit to the largest cluster. For example, using $b{=}4$ and $r{=}8$, 87.5\% of word vectors in the embedding block are quantized to 1-bit. We benefit from 1-bit word vectors in terms of inference speed because memory overhead at matrix multiplications of embedding blocks is potentially minimized. One concern is that 1-bit word vectors may degrade translation performance in a way that is not shown with BLEU score. We address such concerns in Section \ref{sec:qualititaitve-analysis} and demonstrate that 1-bit word vectors do not limit the quantized model's abilities to predict the subsequent tokens.

\subsection{Encoder and Decoder}
\label{sec:encoder-decoder-quant}

Each type of sub-layers in the Transformer yields a wide range of sensitivity to quantization error, and thus, to translation quality drop. Table~\ref{tab:quantization-sesitivity} lists measured BLEU scores with various types of sub-layers quantized into different numbers of quantization bits\footnote{$Emb$, $Enc$, and $Dec$ denote the embedding block, the encoder block, and the decoder block, respectively. $ee$, $ed$, and $dd$ denote the encoder-encoder(encoder self), encoder-decoder, and decoder-decoder(decoder self) attention, respectively. $ffn$ denotes the feed forward sub-layer.}. 
For each type of sub-layers, we carefully select the number of quantization bits such that the model with quantized sub-layers is able to report reasonable degradation in the BLEU score compared to the baseline.

\begin{table}[!t]
\begin{tabular}{cccccc}
\toprule
\multirow{2}{*}{Layer} & \multicolumn{4}{c}{\# of bits} & \multirow{2}{*}{\begin{tabular}[c]{@{}c@{}}Avg.\\ Deg.\end{tabular}} \\ \cline{2-5}
 & 4 & 3 & 2 & 1 &  \\ \midrule
\boldmath$Emb$ & 22.9 & 22.4 & 19.0 & 1.0 & -9.1 \\ \midrule
\boldmath$Enc$ & 24.6 & 24.0 & 20.6 & 1.3 & -7.8 \\
$Enc_{ee}$ & 25.3 & 24.8 & 23.7 & 13.6 & -3.6 \\
$Enc_{ffn}$ & 24.9 & 24.8 & 23.1 & 4.3 & -6.2 \\ \midrule
\boldmath$Dec$ & 24.7 & 23.6 & 11.1 & 0.1 & -10.6 \\
$Dec_{dd}$ & 25.2 & 25.1 & 24.8 & 17.8 & -2.2 \\
$Dec_{ed}$ & 25.1 & 24.7 & 20.6 & 2.0 & -7.3 \\
$Dec_{ffn}$ & 25.0 & 24.9 & 24.4 & 17.6 & -2.5 \\ \bottomrule
\end{tabular}
\caption{\label{tab:quantization-sesitivity} BLEU measurements from applying quantization to each block and to a type of sub-layers in En2De base model (25.4 BLEU in full precision)  without retraining. Avg. Deg. denotes the average BLEU degradation from quantizing each block or a type of sub-layers to 4, 3, 2 and 1 bit. Reported scores are measured in detokenized-BLEU (beam=1, newstest2013, sacrebleu) setting explained in Section \ref{sec:implementation}.}
\end{table}

Within the decoder block, $Dec_{ed}$ sub-layers are more sensitive by quantization than the other sub-layers, which is aligned with reports of \citet{are-sixteen-heads-better-than-one}. 
It is interesting that even though the number of parameters in $Dec_{ffn}$ sub-layers is 2$\times$ that of $Dec_{ed}$ sub-layers, BLEU score degradation is greater when $Dec_{ed}$ sub-layers are quantized. Among the sub-layers in the encoder block, $Enc_{ffn}$ sub-layers are more sensitive by quantization than $Enc_{ee}$ sub-layers. Based on such sensitivity analysis, we assign a proper number of quantization bits to each sub-layer in the encoder and decoder blocks.

Another vital aspect to consider is the inference efficiency of quantized Transformer models. As mentioned in Section \ref{sec:background}, the auto-regressive nature of the Transformer's inference limits the amount of parallelism in the decoder forward propagation and induces a memory wall problem during inference. Therefore, in order to enable fast on-device NMT, we assign a lower number of bits to the decoder block compared to the encoder block.

\begin{table*}[t]
\begin{threeparttable}
  \centering
    \begin{tabular}{ccccccccc}
    \toprule
    \multicolumn{2}{c}{Average \# of Bits}   & \multicolumn{3}{c}{BLEU(beam=1)} & \multicolumn{3}{c}{BLEU(beam=4)} & Model \\
\cline{3-8}    Emb, Dec, Enc & Model\tnote{2} & En2De & En2Fr & En2Jp & En2De & En2Fr & En2Jp & Size(MB) \\
    \midrule
    FP baseline & 32.0 & 26.7 & 39.1 & 25.1 & 27.5 & 39.5& 26.3 & 237.8(1.0$\times$) \\ 
    3-bit baseline & 3.0 & 26.0 & 38.0 & 25.3 & 26.9 & 38.6 & 25.9 & 23.7(10.0$\times$) \\     
    2-bit baseline & 2.0 & 23.9 & 35.4 & 22.8 & 24.4 & 36.1 & 23.7 & 15.9(15.0$\times$) \\    
    2-bit Emb. baseline & 23.7 & 26.0 & N/A & N/A & 26.8 & N/A & N/A & 176.1 \\\midrule
    2.5\tnote{1}, FP, FP   & 23.9  & 26.7  & 39.1 & 25.2 & 27.6  & 39.5  & 25.7 & 177.7 \\
    1.3\tnote{1}, FP, FP   & 23.5  & 26.4  & 38.7 & 24.5 & 27.0  & 39.3  & 24.9  & 175.2 \\
    1.1\tnote{1}, FP, FP   & 23.5  & 25.7  & 38.8 & 24.7 & 26.9  & 39.4  & 25.3  & 174.8 \\ \midrule
    2.5, 1.8, FP  & 11.3  & 25.9  & 38.4 & 24.9 & 27.0  & 38.8  & 25.4  & 85.1 \\
    1.3, 1.8, FP           & 11.0  & 25.6  & 38.1 & 24.5 & 26.8  & 38.8  & 25.1  & 82.5 \\
    1.1, 1.8, FP           & 11.0  & 25.1  & 37.5 & 24.6 & 26.3  & 38.6  & 24.8  & 82.2 \\ \midrule
    2.5, 1.8, 3.7\tnote{3}          & 2.6   & 26.2  & 38.6 & 25.3 & 27.1  & 39.2  & 26.1  & 20.2(11.8$\times$) \\
    1.3, 1.8, 3.7\tnote{3}          & 2.2   & 25.6  & 38.3 & 24.7 & 25.9  & 38.9  & 25.5  & 17.6(13.5$\times$) \\
    1.1, 1.8, 3.7\tnote{3}          & 2.2   & 25.3  & 38.3 & 24.6 & 26.0  & 39.0  & 25.3  & 17.2(13.8$\times$)\\
    \bottomrule
    \end{tabular}
    \begin{tablenotes}
        \footnotesize
        \item[1] For 2.5, 1.3 and 1.1-bit embeddings, Algorithm \ref{alg:embedding} with $b{=}4, r{=}1,4,8$ is applied respectively.
        \item[2] Model column lists the average number of bits in each model.
        \item[3] Average scores over 3 retraining runs are reported for the last retraining phases.
    \end{tablenotes}
  \caption{\label{tab:results-hw-indep}Tokenized-BLEU (beam${\in}\{1,4\}$, newstest2014, multi-bleu) and compression ratio of baseline models and quantized models using proposed quantization strategy. We report BLEU and model size for each retraining phase. All model parameters are included in the reported model size and compression ratio.}
\end{threeparttable}
\end{table*}

\section{Experiments}
\label{sec:results}

\subsection{Quantization Details}
\label{sec:quantization-process}

Before we present our compression results, we describe our quantization method and retraining algorithm in detail.

\paragraph{Methodology} To quantize weights in the Transformer with high performance during retraining, we adopt the Greedy approximation algorithm introduced in \cite{greedy-binary-code-quant} due to its computational simplicity. In our experiments, we first train the base configuration of the Transformer. Next, we retrain the full precision parameters\footnote{In our experiments, full precision parameters are represented by the standard IEEE 32-bit floating-point data format.} while periodically quantizing model parameters to retain the translation quality. For retraining, we adopt Non-Regularization period ($pNR$) as a way to control regularization strength while the best period is empirically obtained \cite{pnr}. Variable $pNR$ is investigated for our retraining, which denotes the number of mini-batch updates before the quantization is performed. For example for $pNR{=}1000$, we first apply quantization to target Transformer weights, and perform 1000 steps of retraining before quantizing the weights again (i.e, the quantization procedure is periodically executed in an interval of 1000 steps during retraining.). The advantage of adopting $pNR$ is reduced retraining time, as computation overheads induced by quantization are divided by $pNR$.

\paragraph{Retraining Details} Our quantization baselines are retrained warm-starting from our full precision baseline. Note that during the retraining, quantization is applied to all layers of the Transformer model every $pNR$ steps where $pNR{=}2000$. Quantization baselines are retrained for 400k steps by using 4$\times$V100 GPUs taking around 1.7 days. 
Our quantized models are retrained over 3 phases in the order of embedding, decoder, and encoder block; each phase warm-starts from the previous phase. Note that in each phase, compressed blocks of previous phases are also targeted for quantization. For each phase, we use $pNR{=}1000$. We train our quantized models for 300k steps/phase and full retraining time is around 3.8 days with 4$\times$V100 GPUs. The reasoning behind the choices of the $pNR$ values and the number of retraining steps is further supported in Appendix \ref{sec:retraining}

\paragraph{Quantized Parameters} Our quantization strategy targets weight matrices that incur heavy matrix multiplications. Targeted weight matrices account for 99.9\% of the number of parameters in the Transformer architecture and 99.3\% of on-device inference latency (Table \ref{tab:flops-lat}). We quantize each row of $W$ as in Figure \ref{fig:bcq-format}, assuming matrix multiplication is implemented as $W \cdot \mathbf{x}$ where $W$ is a weight matrix of model. We do not quantize bias vectors and layer normalization parameters. These parameters account for only a tiny fraction in terms of the total number of parameters and computation overhead, but it is important to retain these parameters in high precision. 
It is commonly acknowledged that quantization error in a bias vector will act as an overall bias \cite{tf-lite}. Also \citet{intel8bit} points out that layer normalization operations will result in high error with low precision parameters as it includes calculations like division, square and square root.

\subsection{Experimental Settings}
\label{sec:implementation}

\paragraph{Dataset} We test our quantization strategy in 3 different translation directions: English-to-German (En2De), English-to-French (En2Fr), and English-to-Japanese (En2Jp). For En2De and En2Fr, we utilize all of the trainset of WMT2014 and use newstest2013 as devset and newstest2014 as testset \cite{wmt2014}. For En2Jp, we use  KFTT \cite{kftt}, JESC \cite{jesc}, and WIT$^3$ \cite{wit3} corpus. We combine the respective trainsets and devsets. We utilize KFTT testset as our testset. All En2Jp data are detokenized as suggested by \citet{mtnt}. \texttt{sentencepiece 0.1.85} \cite{sentencepiece} is utilized to learn a bi-lingual vocabulary set of size 32768 for each translation direction.
For data statistics and download links, refer to Appendix \ref{sec:app-data}.

\paragraph{Baseline Model} We train the base configuration of the Transformer to be utilized as our full precision reference as well as an initial set of model parameters for our quantization experiments.
Training hyper-parameters are listed in Appendix \ref{sec:app-training}.

\paragraph{BLEU} We report both tokenized-BLEU and detokenized-BLEU scores. We report detokenized-BLEU on devsets using \texttt{sacrebleu} \cite{sacrebleu}. While no tokenization is applied to En2De and En2Fr results and devsets, for En2Jp, \texttt{mecab} \cite{mecab} tokenized results and devsets are utilized. Simple \texttt{sacrebleu} command without additional signatures is used to measure detokenized-BLEU.
For testsets, tokenized-BLEU scores are reported. Tokenizers employed are \texttt{moses} \cite{moses} tokenizer\footnote{ {https://github.com/moses-smt/mosesdecoder}} for En2De and En2Fr and \texttt{mecab} \cite{mecab} tokenizer for En2Jp. On the tokenized results and testsets, \texttt{multi-bleu.perl} script in \texttt{moses} is used to measure the tokenized-BLEU score. Note that in each experiment, we report testset's BLEU score using the model parameters that describe the highest BLEU score on devset.

\subsection{Results}
\label{sec:experiments}

We compare our quantization strategy to our full precision (FP) baseline and quantization baselines in terms of translation quality and inference efficiency. Note that for the 2-bit baselines and 3-bit baselines, we respectively assign quantization bits of 2 and 3 to all Transformer parameters, and as for the 2-bit Emb. baseline, we assign 2 quantization bits to all word vectors in embedding block. Our quantized models are notated as (average \# bits in an embedding parameter, average \# bits in a decoder parameter, average \# bits in an encoder parameter).

\paragraph{Translation Quality} In Table \ref{tab:results-hw-indep}, we present translation quality in terms of BLEU scores measured at each phase of the proposed quantization strategy. First, we experiment our embedding quantization method with retraining. Experimental results show that Transformer model with 1.1-bit embedding (1.1, FP, FP) exhibits comparable performance as much as 2-bit Emb. baseline. Furthermore, our experiments with 1.3-bit embedding (1.3, FP, FP) and 1.1-bit embedding verify that a substantially large number of word vectors can be quantized into 1-bit within reasonable BLEU score degradation. 

\begin{table}[t]
  \centering
    \begin{tabular}{cccc}
    \toprule
    Average \# of Bits     & Peak & Avg \\
    Emb, Dec, Enc & MEM.(MB) & Lat.(ms) \\
    \midrule
    FP baseline      & 247.7 & 708.9 \\
    3-bit baseline         & 34.5  & 301.0 \\
    2-bit baseline         & 24.5  & 235.9 \\
    \midrule
    2.5, FP, FP     & 188.3 & 464.3 \\
    2.5, 1.8, FP    & 94.5  & 201.4 \\
    2.5, 1.8, 3.7   & 29.8  & 200.7 \\
    \bottomrule
    \end{tabular}
  \caption{\label{tab:results-hw-dep} Inference latency of our quantized En2De model on a Galaxy N10+. Avg. Lat denotes average latency for translation of an input sequence. All measurements are averaged over 3 runs of translating first 300 sequences in newstest2013. Refer to Appendix \ref{sec:more-details} for measurement and implementation details.
  }
\end{table}

We further quantize Transformer by applying quantization to the decoder block. We study how sensitive each sub-layer is by quantization toward translation quality, and we assign the number of bits for each sub-layer accordingly. Each type of sub-layers in the decoder block are assigned 2, 3, and 1 bits to  $Dec_{dd}$, $Dec_{ed}$, and $Dec_{ffn}$ respectively. In this case, the average of quantization bits for the decoder block is 1.8. For (2.5, 1.8, FP) model, considering that we quantize the embedding and decoder blocks, which account for large number of parameters (69.0\%), into the average of under 3-bit, BLEU score degradation is moderate (within -1 BLEU from the FP baseline).

As we mentioned in Section \ref{sec:transformer}, computations for encoder can be easily parallelizable, and thus, we assign slightly higher number of bits to the encoder block. We can improve quantization result of encoder block to be 3.7-bits per weight by assigning 3 bits to $Enc_{ee}$ sub-layers and 4 bits to more sensitive $Enc_{ffn}$ sub-layers. It is interesting that (2.5, 1.8, 3.7) models in various directions show higher BLEU score than (2.5, 1.8, FP) models which are of previous retraining phases with higher number of bits to represent the models.
\begin{table}[t]
\begin{tabular}{ccc}
\toprule
Block     & FLOPs         & Latency(ms) \\ \midrule
Encoder   & 0.52G(20.8\%)  & 36.4(4.4\%)            \\
Decoder   & 1.49G(59.2\%) & 411.1(49.8\%)            \\
Embedding & 0.50G(20.0\%)  & 372.4(45.1\%)         \\ \hline
Total     & 2.52G         & 825.1            \\ \bottomrule
\end{tabular}
\caption{\label{tab:flops-lat}FLOPs and on-device latency required for translation. Decoder-side activation caching is used. Latency is averaged over 100 translation runs on a Galaxy N10+. A translation run denotes an example translation with 30 words input and output sequences. Note that 300 sequences of newstest2013 consist of 27.6 words in average.}
\end{table}
Our 2.6-bit Transformer models (2.5, 1.8, 3.7) attain 11.8$\times$ model compression ratio with reasonable -0.5 BLEU or less in 3 different translation directions. Our quantized models outperform the 3-bit baselines in both BLEU score and model compression ratio.

\begin{table*}[t]
  \centering
    \begin{tabularx}{\textwidth}{X}
    \toprule
    \textcolor{gray}{\textit{Source}}\\
    Linda Gray, die die Rolle seiner Ehefrau in der Original- und Folgeserie spielte, war bei Hagman, als er im Krankenhaus von Dallas starb, sagte ihr Publizist Jeffrey Lane. \\
    \hline
    \textcolor{gray}{\textit{Reference}}\\
    Linda Gray, who played his wife in the original series  and  the  sequel,  was  with  Hagman  when he died in a hospital in Dallas, said her publicist, Jeffrey Lane. \\
    \textcolor{gray}{\textit{Generated (full-precision model, beam=4)}} \\
    Lind\underline{a} Gra\underline{y}, who played the role of his wife \underline{in} the original and subsequent \underline{series}, was with Hag\underline{man} when he died at Dal\underline{las} \underline{hospital}\underline{,} said her journalist Jeff\underline{re}y Lan\underline{e}. \\
    \textcolor{gray}{\textit{Generated (model with embedding quantized to 1.1 bit, beam=4)}} \\
    \textbf{Lind}\underline{a} \textbf{Gra}\underline{y}, who played the role of his \textbf{wife} \underline{in} the original and \textbf{subsequent} \underline{series}, was with \textbf{Hag}\underline{man} when he died in \textbf{Dal}\underline{\textbf{las}} \underline{\textbf{hospital}}\underline{,} said her publicist \textbf{Jeff}\underline{re}y \textbf{Lan}\underline{e}. \\
    \bottomrule
    \end{tabularx}
\caption{\label{tab:qual-analysis}A De2En translation sample from a FP model and a (1.1, FP, FP) model. Detokenized-BLEU (beam=1, newstest2013, sacrebleu) for each of the models are 30.5 and 30.4. Words with 1-bit quantization are in \textbf{bold letters}. One word with 1-bit quantization is followed by an underlined word. For both full-precision model and quantized model, underlined words are identical.}
\end{table*}

\begin{table}[t]
\begin{tabular}{cccc}
\toprule
\multirow{2}{*}{Method} & \multicolumn{2}{c}{BLEU} & \multirow{2}{*}{Comp.} \\ \cline{2-3}
 & \multicolumn{1}{l}{En2De} & En2Fr &  \\ \midrule
\citeauthor{transformer} & 27.3 & 38.1 & 1.0$\times$ \\
\citeauthor{intel8bit} - 8bit & 27.3 & - & $\leq$4.0$\times$ \\
\citeauthor{fully-quantized-transformer} - 8bit & 27.6 & 39.9 & 3.9$\times$ \\
\citeauthor{fully-quantized-transformer} - 4bit & 18.3 & 1.6 & 7.7$\times$ \\
Ours - 2.6 bit & 27.1 & 38.0 & 11.8$\times$ \\ \bottomrule
\end{tabular}
\caption{\label{tab:results-comparison}Comparison of our quantization strategy with other quantization methods. Comp. denotes compression ratio in terms of model size.}
\end{table}

\paragraph{Inference Speed Up} Let us discuss implementation issues regarding Transformer inference operations for on-device deployment. Measurements of the inference latency and the peak memory size on a mobile device is presented in Table \ref{tab:results-hw-dep}. Our 2.6-bit quantized model (with (2.5, 1.8, 3.7) configuration) achieves 3.5$\times$ speed up compared to the FP baseline. Interestingly, our (2.5, 1.8, FP) model with the average of 11.3-bit outperforms the 2-bit baseline in terms of inference speed. In other words, as for inference speed up, addressing memory wall problems may be of higher priority rather than attaining a low number of quantization bits. 

For each block, Table \ref{tab:flops-lat} shows the number of FLOPs and on-device inference latency. The decoder block demands higher FLOPs than the encoder block (3$\times$), and therefore, employs even higher ratio of on-device inference latency than the encoder block (11$\times$).
Note that while the embedding block requires an amount of FLOPs to be comparable to that of the encoder block, it causes 11$\times$ more inference time than the encoder block. This experiment shows that it is essential to address memory inefficiency for fast on-device deployment of the Transformer.

\paragraph{Comparison} Finally, in Table \ref{tab:results-comparison}, we compare our quantization strategy to previous Transformer quantization methods. All listed methods show results on quantized models based on Transformer base configuration with WMT14 trainsets and report tokenized-BLEU on newstest2014 with exception of \citet{intel8bit} lacking specific BLEU scoring method. Our work outperforms previous quantization studies in terms of compression ratio and achieves reasonable translation quality in terms of BLEU as compared to reported BLEU of full precision models. \citet{intel8bit} reports speed up but it is not directly comparable because of the difference in inference settings (e.g. device used, decoding method, etc.) and other studies do not mention speed up.

\subsection{Qualitative Analysis}
\label{sec:qualititaitve-analysis}

In our strategy, after a large portion of word vectors are quantized by using 1 bit, translation quality degradation may occur even if BLEU does not capture such degradation. Correspondingly, as an attempt to empirically assess the quality of generated translation results with 1-bit quantized word vectors, we investigate how a decoder block predicts the next word. In Table~\ref{tab:qual-analysis}, we present translation examples generated by models with full precision embedding block or with quantized embedding block. Comparing full precision model and quantized model, we observe that for each word with 1-bit quantization, a decoder block generates the same next word (underlined in Table~\ref{tab:qual-analysis}). We present more examples in Appendix \ref{sec:more-qual}. As such, qualitative analysis suggests that our quantization would not noticeably degrade the prediction capability of a decoder even when an input vector is 1-bit quantized.

\section{Related Work}
\label{sec:Transformer-comp}


Previous researches proposed various model compression techniques to reduce the size of Transformer models.  \citet{state-of-sparsity} apply pruning \cite{pruning} to eliminate redundant weights of Transformer and report that higher pruning rates lead to greater BLEU score degradation. As for pruning, achieving inference speed up is more challenging because unstructured pruning method is associated with irregular data formats, and hence, low parallelism \cite{structured-compression}.

Uniform quantization for Transformer is explored within reasonable degradation in BLEU score at INT8, while BLEU score can be severely damaged at low bit-precision such as INT4 \cite{fully-quantized-transformer}. In order to exploit efficient integer arithmetic units with uniformly quantized models, activations need to be quantized as well \cite{tf-lite}. Furthermore, probability mapping operations in Transformer, such as layer norm. and softmax, could exhibit significant amount of error in computational results with low precision data type \cite{intel8bit}.


\section{Conclusion}
\label{sec:conclusion}

In this work, we analyze each block and sub-layer of the Transformer and propose an extremely low-bit quantization strategy for Transformer architecture. Our 2.6-bit quantized Transformer model achieves 11.8$\times$ model compression ratio with reasonable -0.5 BLEU. We also achieve the compression ratio of 8.3$\times$ in memory footprints and 3.5$\times$ speed up on a mobile device (Galaxy N10+).

\bibliography{emnlp2020}
\bibliographystyle{acl_natbib}

\newpage
\clearpage
\appendix

\section{Experiment Details}
\label{sec:more-details}

\subsection{Data}
\label{sec:app-data}

Of data we use, WMT2014 data \cite{wmt2014} includes: Europarl v7 \cite{europarl}, Multi-UN corpus \cite{multiun}, News commentary corpus Giga French-English corpus provided by OPUS \cite{OPUS}, and data provided by CommonCrawl foundation\footnote{ {https://commoncrawl.org/}}. Statistics of data is represented in Table \ref{tab:data-stat}. Data used for En2De and En2Fr can be found at  {https://www.statmt.org/wmt14/translation-task.html}. Data used for En2Jp can be found at: KFTT ( {http://www.phontron.com/kftt/}), WIT$^3$ ( {https://wit3.fbk.eu/mt.php?release=2017-01-trnted}), and JESC ( {https://nlp.stanford.edu/projects/jesc/}).

\begin{table}[h]
\begin{center}
\begin{tabular}{cccc}
\toprule
Translate & \multicolumn{3}{c}{\# of Sequences} \\ \cmidrule{2-4} 
Direction & Train & Dev & Test \\ \midrule
En2De& 4.5M & 3000 & 3003 \\
En2Fr & 40.8M & 3000 & 3003 \\
En2Jp & 3.9M & 4451 & 1160 \\ \bottomrule
\end{tabular}

\end{center}
\caption{\label{tab:data-stat}Statistics of data used for each translation direction.}
\end{table}

\subsection{Model}
\label{sec:app-model}
All models follow the base configuration of Transformer architecture composed of 60.9 million parameters \cite{transformer}.

\subsection{Training}
\label{sec:app-training} Our training and retraining implementation is based on \texttt{tensor2tensor} 1.12's implementation of Transformer and utilizes \texttt{tensorflow 1.14} \cite{tensorflow} modules. All training hyperparameters exactly follow \texttt{transformer\_base} configuration of the code. We use 4$\times$V100 GPUs for all training and retraining, and for each training step, a mini-batch of approximately 8,000 input words and 8,000 target words is used per GPU. Training of a full precision baseline model takes around 1.7 days. Adam optimizer \cite{adam} with $\beta_1=0.9, \beta_2=0.999, \epsilon=10^{-9}$ is used and we adopt Noam learning rate scheme of \citet{transformer} using same suggested hyperparameters. Baseline models are trained for 400,000 training steps and we select models that have the highest BLEU score on devset to report as our full precision baseline and to warm start from in our retraining for quantization.

\subsection{Retraining}
\label{sec:retraining} For retraining, we experiment with $pNR \in \{1, 10, 100, 500, 1000, 2000, 4000\}$. With $pNR \in \{1, 10, 100\}$, our retraining experiments resulted in divergence. We find that for a retraining phase where we quantize all blocks of Transformer, $pNR=2000$ is the most effective in attaining a higher BLEU score with quantized model. And for a retraining phase in 3-phase retraining, where we quantize a block in Transformer, $pNR=1000$ is the most effective. Hence, we set $pNR=2000$ for retraining of quantization baselines, and for experiments where we quantize and retrain each block in Transformer at a time, we set $pNR=1000$. While the choice for the value of $pNR$ is made in empirical manner, it should be noted that in our tests, regardless of the number of quantization bits or other design choices, the choice of $pNR$ value between 1000 and 2000 did not result in high variance on translation quality. 

Our learning rate ($lr$) schedule is similar to the Noam schedule suggested in \cite{transformer}, but replaced the warm-up stage with a constant $lr$ stage as in Eq. \ref{eq:lr}:

\begin{align}
\label{eq:lr}
lr = c_{lr} \cdot d_{model}^{0.5} \cdot min(step^{-0.5}, steps_{peak}^{-0.5})
\end{align}

$step$ is incremented by 1 with each mini-batch update and reset to 0 at each retraining phase. We use $c_{lr}=3$ for all retraining. This scheme results in higher overall learning rate than what we use in our full precision baseline training, which follows the heuristics that large enough learning rate is required to find the best local minima with quantization constraint applied.

For single-phase retraining, we train up to 400,000 steps. Based on BLEU score on devset, single-phase retraining seems to reach convergence at around 300,000 steps. As for 3-phase retraining, we train for 300,000 steps respectively. We found 300,000 steps ample for a retraining phase to reach convergence judging from the reported BLEU scores on the validation set. In the 3-phase retraining, we first retrain and quantize embedding then embedding + decoder and finally all blocks of Transformer. For each phase of retraining, we take a model that reports the highest detokenized-BLEU score on devset. Retraining hyperparameters that are not stated follow corresponding hyperparameters of full precision model training Additionally, we attempt another variant of 3-phase retraining where we target only a single Transformer block at each phase and stop gradients on previously targeted Transformer blocks. However, this method of retraining results mostly in moderately lower BLEU score compared to our current 3-phase retraining method.

\subsection{On-Device Inference}
\label{sec:app-on-device}
On-device inference is implemented with Eigen 3.7 \cite{eigen} for full precision computation and BiQGEMM \cite{biqgemm} for computation with quantized weights. With BiQGEMM, the value of redundant intermediate computation that occurs in matrix multiplication of quantized weights is pre-computed and stored to be reused, which is promising in reduction of memory overhead. Each $B$ value is represented with a single bit in memory where 0 denotes -1 and 1 denotes +1, and in our implementation bits are packed into 32-bit integer which is directly used at inference. We follow BiQGEMM in our implementation of quantized inference. In our implementation, we implement decoder-side activation caching following \texttt{tensor2tensor}'s implementation of Transformer. We measure on-device latency with a \texttt{<chrono>} implementation of C++14 and memory usage with \texttt{adb}\footnote{ {https://developer.android.com/studio/command-line/adb}}. Unless otherwise specified, both latency and memory usage are measured while translating the first 300 sequences of En2De testset over 3 translation runs. Additional statistics regarding inference latency and memory of quantized models are available in Table \ref{tab:results-hw-dep-details}.

\begin{table*}[htbp]
  \centering
    \begin{tabular}{ccccccccc}
    \toprule
    Average \# of Bits & \multicolumn{4}{c}{Latency contribution(\%)} & Avg. & Peak & \multicolumn{2}{c}{Avg. \# of Words} \\
\cmidrule{2-5}\cmidrule{8-9}    Emb, Enc, Dec & Emb   & Enc   & Dec   & Other & Lat(ms) & MEM(MB) & Input & Output \\
    \midrule
    FP baseline & 44.4\% & 4.9\% & 50.1\% & 0.6\% & 708.0 & 247.7 & 27.6  & 27.9 \\
    3-bit baseline & 51.3\% & 8.7\% & 38.3\% & 1.8\% & 301.0 & 34.5  & 27.6  & 28.0 \\
    2-bit baseline & 45.9\% & 9.5\% & 42.2\% & 2.3\% & 235.9 & 24.5  & 27.6  & 28.0 \\
    \midrule
    2.5, FP, FP & 14.4\% & 7.0\% & 77.3\% & 1.2\% & 464.3 & 188.3 & 27.6  & 28.4 \\
    2.5, 1.8, FP & 34.6\% & 16.3\% & 46.2\% & 2.8\% & 201.4 & 94.5  & 27.6  & 28.4 \\
    2.5, 1.8, 3.7 & 34.4\% & 16.5\% & 46.5\% & 2.7\% & 200.7 & 29.8  & 27.6  & 27.7 \\
    \bottomrule
    \end{tabular}%
  \caption{\label{tab:results-hw-dep-details} Additional statistics regarding reported measurements of Table \ref{tab:results-hw-dep}.
  }
\end{table*}%

\section{Validation Score}
\label{sec:results-eval}

We report the validation scores (detokenized-BLEU scores on devset) of experimented models in Table \ref{tab:results-eval}.

\begin{table*}[t]
\centering
\begin{tabular}{ccccc}
\toprule
\multicolumn{2}{c}{Average \# of Bits} & \multicolumn{3}{c}{Validation BLEU(beam=1)} \\ \cmidrule{3-5} 
Emb, Dec, Enc & Model & En2De & En2Fr & En2Jp \\ \midrule
FP baseline & 32.0 & 25.4 & 31.4 & 18.7 \\
3-bit baseline & 3.0 & 25.3 & 30.1 & 18.0 \\
2-bit baseline & 2.0 & 23.9 & 28.6 & 16.8 \\
2-bit baseline(Emb) & 23.7 & 25.1 & N/A & N/A \\ \midrule
2.5, FP, FP & 23.9 & 25.6 & 31.2 & 18.5 \\
1.3, FP, FP & 23.5 & 25.3 & 31.0 & 17.5 \\
1.1, FP, FP & 23.5 & 25.2 & 31.0 & 17.9 \\ \midrule
2.5, 1.8, FP & 11.3 & 25.2 & 30.6 & 18.1 \\
1.3, 1.8, FP & 11.0 & 24.6 & 30.4 & 17.7 \\
1.3, 1.8, FP & 11.0 & 24.4 & 30.4 & 17.2 \\ \midrule
2.5, 1.8, 3.7 & 2.6 & 25.1 & 30.9 & 18.4 \\
1.3, 1.8, 3.7 & 2.2 & 24.9 & 30.6 & 17.7 \\
1.1, 1.8, 3.7 & 2.2 & 24.4 & 30.5 & 17.6 \\ \bottomrule
\end{tabular}
\caption{\label{tab:results-eval}BLEU score on devset of baseline models and quantized models. We report detokenized-BLEU (beam=1, newstest2013, sacrebleu) for En2De, En2Fr as suggested in in Section \ref{sec:implementation}. For En2Jp, outputs and references are tokenized with \texttt{mecab} then measured with \texttt{sacrebleu}.}
\end{table*}

\section{Sequences Generated with 1-bit Words}
\label{sec:more-qual}

\begin{table*}[t]
  \centering
    \begin{tabularx}{\textwidth}{X}
    \toprule
        \textcolor{gray}{\textit{Source 1}} \\
 Im vergangenen Jahr gingen beim CTMO mehr als 1,4 Millionen Anträge auf Markenschutz ein, fast ein Drittel mehr als 2010.
\\
    \midrule
    \textcolor{gray}{\textit{Reference 1}} \\
In the past year, more than 1.4 million applications for trademark protection were submitted to the CTMO, almost one third more than in 2010.\\
    \midrule
    \textcolor{gray}{\textit{Generated 1 (full-precision model, beam=4)}} \\
    Last year, more than 1.4 \underline{million} applications for trademark \underline{protection} were received at the CTMO, almost one third more than in 2010.
\\
    \midrule
    \textcolor{gray}{\textit{Generated 1 (model with embedding quantized to 1.1 bit, beam=4)}} \\
    Last year CTMO received more than \textbf{1.4} \underline{million} \textbf{trademark} \underline{protection} applications, almost a third more than in \textbf{2010.}
     \\
    \bottomrule
    \toprule
    \textcolor{gray}{\textit{Source 2}} \\
    Der derzeitige Premierminister Israels, der Falke Netanjahu, ist ein typisches Beispiel eines faschismusanfälligen, den internationalen Bankern loyal ergebenen Politikers, der alles dafür tut, um einen Krieg mit dem Iran zu entfachen, welcher sich angesichts der Mitgliedschaft Irans in der Schanghaier Organisation für Zusammenarbeit (China, Indien, Russland, Pakistan...), rasch zu einem globalen Konflikt ausweiten könnte, und bei dem es wegen der Kontrolle Irans über die nur 2 Meilen breite Straße von Hormus, über die 20\% der weltweiten Erdöllieferungen laufen, zu einer Zerstörung der Weltwirtschaft kommen könnte. \\
    \midrule
    \textcolor{gray}{\textit{Reference 2}} \\
    Israel's current prime minister, Netanyahu 'the hawk', is a typical example of a fascist politician, loyal to the international bankers, who does everything to instigate war with Iran, which would, due to its membership in the Shanghai Cooperation Organisation (China, India, Russia, Pakistan, ...) lead to a greater threat of global conflict, and through its control of the Hormuz Strait, where 20\% of the world's oil must sail (the channel is only 2 miles wide), to the destruction of the world's economy. \\
    \midrule
    \textcolor{gray}{\textit{Generated 2 (full-precision model, beam=4)}} \\
    The current Prime Minister of Israel, the Falk \underline{Net}\underline{an}yahu, is a typical \underline{example} of a fasc\underline{ism}-prone politician \underline{loyal} to international bankers \underline{who} is doing everything possible to spark a war with Iran, which, given Iran's membership \underline{of} the Shanghai \underline{Cooperation} \underline{Organisation} (China\underline{,} India, Russia, Pakistan...)\underline{,} could rapidly spread to a global conflict, and could lead to the destruction \underline{of} the world economy because of Iran's control of the only 2-\underline{mile}\underline{-}wide \underline{Stra}\underline{it} of Hor\underline{mus}\underline{,} which accounts \underline{for} 20\% \underline{of} world oil supplies\underline{.} \\
    \midrule
    \textcolor{gray}{\textit{Generated 2 (model with embedding quantized to 1.1 bit, beam=4)}} \\
    Israel's current \textbf{prime} \underline{\textbf{minister}}\underline{,} \textbf{Falk}\underline{e} \textbf{Net}\underline{an}yahu, is a \textbf{typical} \underline{example} of a fa\textbf{sc}\underline{ism}-prone \textbf{politician} \underline{\textbf{loyal}} to international \textbf{bankers} \underline{who} is doing all he can to \textbf{trigger} \underline{a} war with Iran, which, with Iran's \textbf{membership} \underline{of} the \textbf{Shanghai} \underline{\textbf{Cooperation}} \underline{Organisation} (\textbf{China}\underline{,} India, Russia, Pakistan\textbf{...)}\underline{,} could \textbf{rapidly} develop into a global conflict and could lead to the \textbf{destruction} \underline{of} the world economy because of Iran's control of the only 2 \textbf{mile}\underline{-}\textbf{wide} \underline{\textbf{Stra}}\underline{it} of \textbf{Hor}\textbf{\underline{mus}}\underline{,} which \textbf{accounts} \underline{for} \textbf{20\%} \underline{of} world oil \textbf{supplies}\underline{.} \\
    \bottomrule 
    \end{tabularx}
\caption{\label{tab:more-qual-analysis}De2En translation samples from full-precision model and model with embedding block quantized to 1.1-bit ($b=4, r=8$) with Algorithm \ref{alg:embedding} (1.1, FP, FP). Same models as Table \ref{tab:qual-analysis} is utilized.}
\end{table*}

In Table \ref{tab:more-qual-analysis}, we present actual translation results from full precision embedding block and quantized embedding block. In the first example, 2 out of 2 \underline{words that follow 1-bit words} are equal to their \underline{positional equivalents} in the output sequence generated with the full precision model. In the second example, 19 out of 21 matches.

\end{document}